\documentclass[runningheads]{llncs}
\usepackage{graphicx}
\usepackage{amsmath,amssymb} 
\usepackage{color}
\usepackage[width=122mm,left=12mm,paperwidth=146mm,height=193mm,top=12mm,paperheight=217mm]{geometry}
\begin{document}
\pagestyle{headings}
\mainmatter
\def\ECCV16SubNumber{20}  

\title{Deconvolutional Feature Stacking for Weakly-Supervised Semantic Segmentation}


\author{Hyo-Eun Kim and Sangheum Hwang}
\institute{Lunit Inc., Seoul, South Korea\newline\{hekim, shwang\}@lunit.io }

\maketitle

\begin{abstract}
A weakly-supervised semantic segmentation framework with a tied deconvolutional neural network is presented. Each deconvolution layer in the framework consists of unpooling and deconvolution operations. `Unpooling' upsamples the input feature map based on unpooling switches defined by corresponding convolution layer's pooling operation. `Deconvolution' convolves the input unpooled features by using convolutional weights tied with the corresponding convolution layer's convolution operation. The unpooling-deconvolution combination helps to eliminate less discriminative features in a feature extraction stage, since output features of the deconvolution layer are reconstructed from the most discriminative unpooled features instead of the raw one. This results in reduction of false positives in a pixel-level inference stage.  All the feature maps restored from the entire deconvolution layers can constitute a rich discriminative feature set according to different abstraction levels. Those features are stacked to be selectively used for generating class-specific activation maps. Under the weak supervision (image-level labels), the proposed framework shows promising results on lesion segmentation in medical images (chest X-rays) and achieves state-of-the-art performance on the PASCAL VOC segmentation dataset in the same experimental condition.
\end{abstract}

\section{Introduction}
Deep neural networks have recently been achieved breakthroughs in several domains such as computer vision~\cite{b1_Deng2009imagenet,b2_Krizhevsky2012cls_alexnet,b3_simonyan2014cls_vgg}, speech recognition~\cite{a1_hinton2012speech,a2_dahl2012speech}, and natural language processing~\cite{a3_collobert2008nlp,a4_cho2014nlp}. Especially in computer vision, deep convolutional neural networks (CNNs) are actively applied to object recognition tasks like object classification~\cite{b2_Krizhevsky2012cls_alexnet,b3_simonyan2014cls_vgg}, detection~\cite{b4_bell2015det_insideoutside,b5_Erhan2014det,b6_Oquab2015det_free}, and semantic segmentation~\cite{c1_long2014seg_fully,d2_hong2015seg_semi}. Given an input image, semantic segmentation task should finely estimate pixel-level class labels while object classification just classifies its image-level category, so features discriminating details of target objects as well as semantic information of the entire image should be well defined in a training stage. 

Among the semantic segmentation tasks, semi- or weakly-supervised approaches under weak supervision such as bounding-box annotations~\cite{d1_dai2015seg_semi}, a limited number of segmentation annotations~\cite{d2_hong2015seg_semi,d3_papandreou2015seg_semi}, or image-level labels~\cite{e1_Pinheiro2015seg_weakly,e3_pathak2015seg_weakly,e4_hong2015seg_weakly,e5_zhang2015seg_weakly},  are preferred in real applications, since pixel-level labelling for fully-supervised semantic segmentation~\cite{c1_long2014seg_fully,c2_chen2014seg_fully,c3_hariharan2014seg_fully,c4_hariharan2014seg_fully,c5_mostajabi2014seg_fully,c6_noh2015seg_fully,c7_dai2015seg_fully,c8_pinheiro2015seg_fully} requires heavy annotation efforts compared to the semi- or weakly-supervised counterparts. Especially in weakly-supervised semantic segmentation, only the image-level labels are available for training so the pixel-level fine-grained inference for detailed shape near boundary of target objects is quite difficult. Thus, most of the weakly-supervised semantic segmentation approaches exploit appropriate pre/post-processing or additional informative supervision (e.g., superpixel~\cite{e5_zhang2015seg_weakly}, extraneous segmentation annotations~\cite{e4_hong2015seg_weakly}, size constraints of region-of-interests (ROIs)~\cite{e3_pathak2015seg_weakly}, and smoothing boundary priors~\cite{e1_Pinheiro2015seg_weakly}). 
In natural images, training under those informative priors enables more accurate pixel-level inference while exploiting spatial coherency between pixels.

The ROI segmentation task becomes more challenging in medical images, since only the trained clinicians who have expertise in corresponding medical domains can annotate pixel-level abnormalities. Furthermore, additional image processing commonly used in natural images cannot guarantee improvement of segmentation performance because of different characteristics of target ROIs (e.g., lesion). In this case, domain-specific pre/post-processing is required, but it also needs domain-specific knowledges and expertise.

In this work, a novel method for weakly-supervised semantic segmentation, \textit{deconvolutional feature stacking}, is proposed. We build a deconvolutional neural network on top of the CNN to reconstruct a rich set of discriminative features from the abstracted features of the top most convolution layer. In a single deconvolution layer, input features are upsampled via unpooling switches defined by the corresponding pooling operation (features are bypassed according to the pooled position, and the rests of the upsampled positions are filled with zero~\cite{g1_zeiler2011deconv,g2_zeiler2014deconv}) and convolved by a deconvolution operator. This helps to suppress less discriminative features in a feature extraction stage, since the deconvolution operator reconstructs detailed features from the most discriminative activations. The convolutional weight of the deconvolution operation is tied with that of corresponding convolution layer. This assists training under weak supervision, because `tied weight' confines the search space under the constraint of tight connection between convolution and deconvolution layers. The restored features from all of the deconvolution layers can constitute a rich feature set according to different sizes of receptive fields, and those features are fully utilized in a pixel-level inference stage.

\begin{figure*}[t]
\begin{center}
\includegraphics[width=\textwidth]{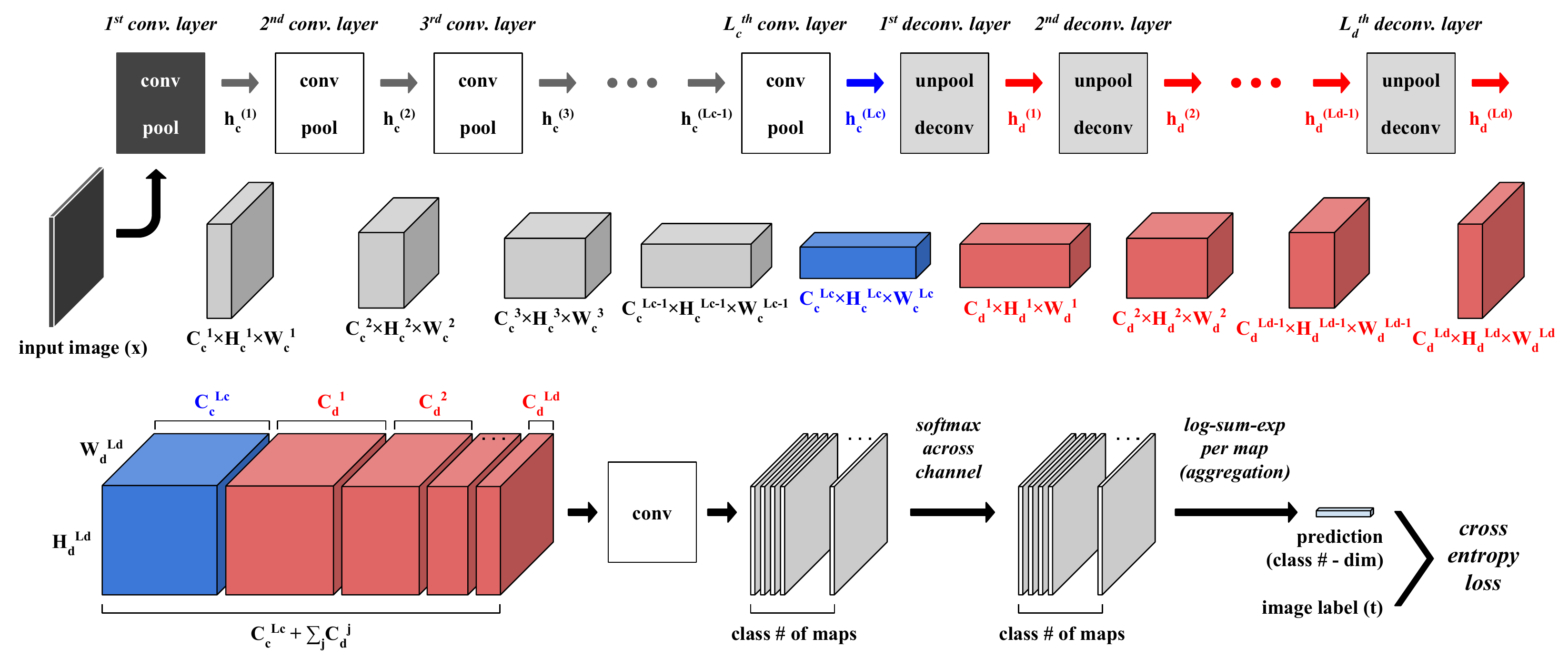}
\caption{Overall architecture of the proposed framework (best viewed in color). Each convolution layer consists of convolutional and pooling operations, and each deconvolution layer consists of unpooling (based on pooling masks of corresponding pooling layers) and deconvolution (convolutional weights tied with corresponding convolution layers) operations. Based on a feature map with the highest abstraction level extracted from the top most convolution layer (blue), details of features are restored using deconvolution layers (red). The entire feature maps are expanded appropriately to be concatenated across channel dimension followed by an additional convolution operation to generate class-specific activation maps. Those maps are softmaxed across channel dimension and aggregated into a single vector to be compared with the image-level label vector, \textbf{t}.}
\vskip -0.2in
\label{fig1:overall_arch}
\end{center}
\end{figure*}

Fig.~\ref{fig1:overall_arch} shows overall architecture of the proposed framework. Features with the highest abstraction level extracted from the last convolution layer is used for building-up details of ROIs. All the feature maps restored from deconvolution layers and the feature map extracted from the last convolution layer are upscaled to be matched with the dimension of the final feature map extracted from the last deconvolution layer. Then, all the feature maps are stacked across channel dimension. The stacked feature map includes from coarse-grained to fine-grained features, so the following convolution layer (bottom of the figure) can selectively extract class-specific key features from the stacked rich feature set. The last convolution layer consists of class number of convolutional filters. Its output maps are softmaxed across channel dimension to assign a single label to each pixel. Since only the image-level labels are available in the weakly-supervised setting, each class-specific map is aggregated into a single value using a global per-map pooling operation. The proposed methodology will be discussed in more detail in Section 3. Our contribution is summarized as below:
\begin{itemize}
\vskip 0.1in
\item[\checkmark] To the best of our knowledge, we present an efficacy of stacking features extracted from a sequence of unpooling-deconvolution operations for the first time in weakly-supervised semantic segmentation. By employing the consecutive deconvolution layers, ROIs with different scales can be covered by a single trainable network.
\vskip 0.05in
\item[\checkmark] We exploit `tied weight' for deconvolutions in order to train the network efficiently under the proposed framework. Especially in small-scale datasets, layer-by-layer training of deconvolution layers helps to localize ROIs in a weakly-supervised setting.
\vskip 0.05in
\item[\checkmark] We experimentally show that the proposed methodology is robust against different image modalities. Without any domain-specific pre/post processing, the proposed framework achieves promising segmentation performances in medical and natural images. 
\vskip 0.1in
\end{itemize}

The rest of this paper is organized as follows. Section 2 introduces related works, and Section 3 describes the detailed methodology. Experimental set-up and results are presented in Section 4, and Section 5 concludes this paper.

\section{Related work} 
Semantic segmentation can be divided into three categories according to its supervision level; fully-supervised, semi-supervised, and weakly-supervised approaches. In fully-supervised semantic segmentation, pixel-level labels are used for training so it is relatively easier to discriminate details of ROIs on an input image~\cite{c1_long2014seg_fully,c2_chen2014seg_fully,c3_hariharan2014seg_fully,c4_hariharan2014seg_fully,c5_mostajabi2014seg_fully,c6_noh2015seg_fully,c7_dai2015seg_fully,c8_pinheiro2015seg_fully}. The semi-supervised semantic segmentation approach is sub-classified into two types according to the type of supervision; bounding box annotations~\cite{d1_dai2015seg_semi} which is useful for multi-scale dataset augmentation or a limited number of segmentation annotations ~\cite{d2_hong2015seg_semi,d3_papandreou2015seg_semi}. Although the fully- or semi-supervised learning for semantic segmentation performs well in real applications, they require heavy annotation efforts in terms of the quality and the amount of annotations.

To overcome the limitations of the fully- or semi-supervised approaches, weakly-supervised semantic segmentation methods trained only with image-level labels are presented recently~\cite{e1_Pinheiro2015seg_weakly,e3_pathak2015seg_weakly,e4_hong2015seg_weakly,e5_zhang2015seg_weakly}. In~\cite{e1_Pinheiro2015seg_weakly}, coarse-grained per-class activation maps are generated from the top convolution layer followed by per-map aggregation (global pooling) using \textit{log-sum-exp}. It is quite similar to~\cite{b6_Oquab2015det_free}, a weakly-supervised approach for object localization, which builds per-class activation maps using image-level labels based on max-pooling for per-map aggregation. A radical difference between those two works is that \cite{e1_Pinheiro2015seg_weakly} uses several segmentation priors on coarse-grained output activation maps in order to reduce false positives for improving segmentation performance. Especially the smoothing prior used in this work is based on the assumption that objects have well defined boundaries and shapes. But in unusual cases like medical images, the assumption is quite ambiguous to be identically applied. Global pooling from the class-specific activation maps such as max-pooling~\cite{b6_Oquab2015det_free} or \textit{log-sum-exp}~\cite{e1_Pinheiro2015seg_weakly} is quite straightforward, so we set this method as a baseline for our work in following sections. 

In~\cite{e3_pathak2015seg_weakly}, training objective of nonlinear deep networks is defined as a linear biconvex optimization model. Based on this model, additional weak supervision such as sizes of background, foreground, or objects can be used as constraints to relax learning target objectives. The constraints used in this work are less informative than pixel-level annotations, but acquisition of those needs additional annotation efforts as fully- or semi-supervised approaches did. 

Weakly-supervised semantic segmentation for noisy images such as wrong or omitted labels is presented in~\cite{e5_zhang2015seg_weakly}. They extract superpixels from input images in order to perform superpixel-level inference. This is based on the assumption that objects have clear boundaries according to spatial coherency between pixels. But, this assumption cannot be guaranteed in medical images, since lesions have different characteristics from the general objects in natural images. 

In~\cite{e4_hong2015seg_weakly}, the authors demonstrate that knowledge is transferable between two different datasets. The trained knowledge on a dataset which has pixel-level segmentation annotations can be exploited for training another network under the dataset only with weak image-level labels. Although the segmentation annotations used for knowledge transfer do not contain categories in the target dataset under the weakly-supervised setting, it can be classified into another type of semi-supervised approaches in terms of using pixel-level segmentation annotations.

We will provide a weakly-supervised semantic segmentation framework which is robust against false positives without additional pre/post processing. It can reconstruct details of ROIs without any types of additional supervision except for the image-level labels. Details of the proposed methodology will be discussed in next section.

\section{Proposed Methodology}
The target task can be interpreted as a multiple-instance learning (MIL)~\cite{f1_maron1998mil,f2_Andrews2002mil}. Under the MIL framework, each pixel is an individual instance and the image is a bag of the instances. Given a bag label (image-level label), instance-level labels (pixel-level labels) should be defined in an inference stage under the condition of MIL; at least one instance is positive if the bag label is positive. In the proposed framework, a rich set of feature maps are extracted from different layers with different abstraction levels in order to classify each instance in a bag correctly. In this section, details of the proposed methodology as well as brief experimental analysis of an effect of the proposed framework will be presented. 

\subsection{Model}
Fig.~\ref{fig1:overall_arch} shows the overall architecture of the proposed framework. The output feature map of the \textit{i}$^{th}$ convolution layer (\textit{i} = 1, 2, ... , $L_c$; ${L_c}$ is the total number of convolution layers) is:
\begin{equation}
\label{eq1:conv_out}
\textbf{h}_{c}^{(i)} = \sigma(\textbf{h}_{c}^{(i-1)} * \textbf{W}_{c}^{(i)} + \textbf{b}_{c}^{(i)})
\end{equation}
\noindent where $\sigma$, *, \textbf{W}$_{c}^{(i)}$, and \textbf{b}$_{c}^{(i)}$ are a nonlinear activation function, a convolutional operator, a convolutional filter weight and a bias of the \textit{i}$^{th}$ convolution layer, respectively. An input feature map of the \textit{1}$^{st}$ convolution layer, \textbf{h}$_{c}^{(0)}$, is an input image \textbf{x} as shown in Fig.~\ref{fig1:overall_arch}. Similarly, the output of the \textit{j}$^{th}$ deconvolution layer (\textit{j} = 1, 2, ... , ${L_d}$; ${L_d}$ is the total number of deconvolution layers) is defined as:
\begin{equation}
\label{eq2:deconv_out}
\textbf{h}_{d}^{(j)} = \sigma(\textbf{h}_{d}^{(j-1)} * \textbf{W}_{d}^{(j)} + \textbf{b}_{d}^{(j)})
\end{equation}
\noindent where \textbf{W}$_{d}^{(j)}$ and \textbf{b}$_{d}^{(j)}$ are a convolutional filter weight and a bias of the \textit{j}$^{th}$ deconvolution layer. In this case, the input of the \textit{1}$^{st}$ deconvolution layer, \textbf{h}$_{d}^{(0)}$, will be the feature map extracted from the last convolution layer, \textbf{h}$_{c}^{(L_c)}$. We use a rectified linear unit (ReLU) for the nonlinearity $\sigma$ in Eq.~(\ref{eq1:conv_out}) and (\ref{eq2:deconv_out}). Pooling and unpooling operations are also included in the convolution and deconvolution layers. `Max-pooling' is used for the pooling operations in convolution layers, and the pooling masks made by the pooling operations are used for unpooling in deconvolution layers. Those two operations are straightforward as explained in Section 1, so it is omitted in following equations for brevity. The experimental analysis of false positive reduction effect caused by the unpooling operation will be described in Section 3.2. 

In a deconvolution stage, each deconvolution layer uses a convolutional weight tied with that of the corresponding convolution layer as below:
\begin{equation}
\label{eq3:tied_weight}
\textbf{W}_{d}^{(j)} = \textbf{W}_{c}^{(L_c+1-j)^T}
\end{equation}
\noindent where \textit{T} is a matrix transpose operation. In a stacked autoencoder framework~\cite{g3_vincent2010sdae}, `tied weight' for an encoder-decoder pair forces the weights of the encoder and the decoder to be symmetric while reconstructing the input from the encoded feature under relatively low-dimensional parameter space. Although our target objective is not the same as unsupervised feature learning of the stacked autoencoder, `tied weight' can be used as an additional constraint for the weakly-supervised learning. Training proceeds to find a good local optimum under the weak supervision while maintaining a tight connection between the encoder (convolution layer) and the decoder (corresponding deconvolution layer) paths under the `tied weight' constraint. In our experiment, training loss in terms of image-level classification is also reduced properly with untied deconvolutions, but performance of ROI localization is significantly worse than the tied one.

Now, we can build a rich feature set from the feature maps generated by the last convolution layer and all of the deconvolution layers as below:
\begin{equation}
\label{eq4:f_concat}
\textbf{f}^{(L_d)} = C\{E\{\textbf{h}_{c}^{(L_c)}\}, E\{\textbf{h}_{d}^{(1)}\}, ... , E\{\textbf{h}_{d}^{(L_d)}\}\}
\end{equation}
\noindent where E$\{\textbf{h}^{(i)}\}$ normalizes the input feature map, \textbf{h}$^{(i)}$, to have zero-mean/unit-variance and repeatedly expands the normalized input feature map in a nearest-neighbor manner with respect to the height and width ratios as follows: 
\begin{equation}
\label{eq5:expand_ratio}
r_{height}^{(i)} = \frac{H^{L_d}}{H^{i}},~~r_{width}^{(i)} = \frac{W^{L_d}}{W^{i}} 
\end{equation}
\noindent where $H^{L_d}$ and $W^{L_d}$ are the height and width of the feature map extracted from the last deconvolution layer, \textbf{h}$_d^{(L_d)}$, and $H^{i}$ and $W^{i}$ are the height and width of the expand target, \textbf{h}$^{(i)}$. C$\{\cdot\}$ operator in Eq.~(\ref{eq4:f_concat}) concatenates the normalized and expanded feature maps across the channel dimension as shown in Fig.~\ref{fig1:overall_arch}.

Based on \textbf{f}$^{(L_d)}$ including from coarse-grained to fine-grained features, class-specific activation maps are generated by an additional convolution operation as below:
\begin{equation}
\label{eq6:maps}
\textbf{h}_{m} = \textbf{f}^{(L_d)} * \textbf{W}_{m} + \textbf{b}_{m} \in \textbf{R}^{K \times H^{L_d} \times W^{L_d}}
\end{equation}
\noindent where \textbf{W}$_{m}$ and \textbf{b}$_{m}$ are a convolutional weight and a bias of the additional convolution operation, and ${K}$ is the number of output activation maps. In a binary classification problem (e.g., abnormality detection in medical images), ${K}$ is two as usual. For the multi-label classification, ${K}$ includes an additional class for `background'. Since pixel-level annotations are not available in a weakly-supervised semantic segmentation, we cannot know the true label for the additional `background' class. So, we assume that all of the training examples include at least one pixel of `background'. This assumption is not correct for some training examples which do not have any `background' pixels. In this case, the positive label for `background' becomes a noisy label. The additional activation map for `background' helps to assign individual pixel to the most probable class under weak supervision. We use a softmax operation, since it suppresses the rest elements of a target vector to put more weight on a specific element. In terms of each pixel component in an image, the target task can be interpreted as a 1-of-\textit{K} classification problem, so we perform a softmax operation across the channel dimension ($\sigma$ in Eq.~(\ref{eq7:LSE})). In a binary classification, element-wise ReLU is used for the nonlinearity.

To be compared with the true image-level label vector (\textbf{t} in Fig.~\ref{fig1:overall_arch}) under the cross-entropy objective function, the final activation maps, $\sigma\left(\textbf{h}_{m}\right)$, are aggregated into a single vector by global per-map pooling (\textit{log-sum-exp}) as below:
\begin{equation}
\label{eq7:LSE}
\textbf{y}_{pred}^{c} = \frac{1}{s} \log \left( 
\frac{\sum_{i,j}\exp \left( s \cdot \sigma \left(\textbf{h}_{m} \right) ^{c}_{i,j} \right)}{H^{L_d} W^{L_d}} 
\right).
\end{equation}
The superscript \textit{c} describes class-\textit{c}, so \textbf{y}$_{pred}^{c}$ is the \textit{c}$^{th}$ element of a final prediction vector. $\sigma \left(\textbf{h}_{m} \right)^{c}_{i,j}$ includes all the within-class activation values of the \textit{c}$^{th}$ activation map. ${s}$ in Eq.~(\ref{eq7:LSE}) is a control parameter for similarity between within-class activations. Smaller $s$ mimics average-pooling while the larger one is used for more max-pooling-like operation.

The cross-entropy loss between a prediction vector \textbf{y}$_{pred}$ and its true label \textbf{t} is used for the final objective function. Specifically, a sum of \textit{K} binary cross-entropy losses and a categorical cross-entropy loss are used for multi-label classification and binary classification, respectively. In binary classification, a softmax operation is performed on \textbf{y}$_{pred}$ before adopting the target objective function (categorical cross-entropy loss). 

\begin{figure}[t]
\begin{center}
\includegraphics[scale=0.3, trim=1cm 0cm 0cm 1cm, clip]{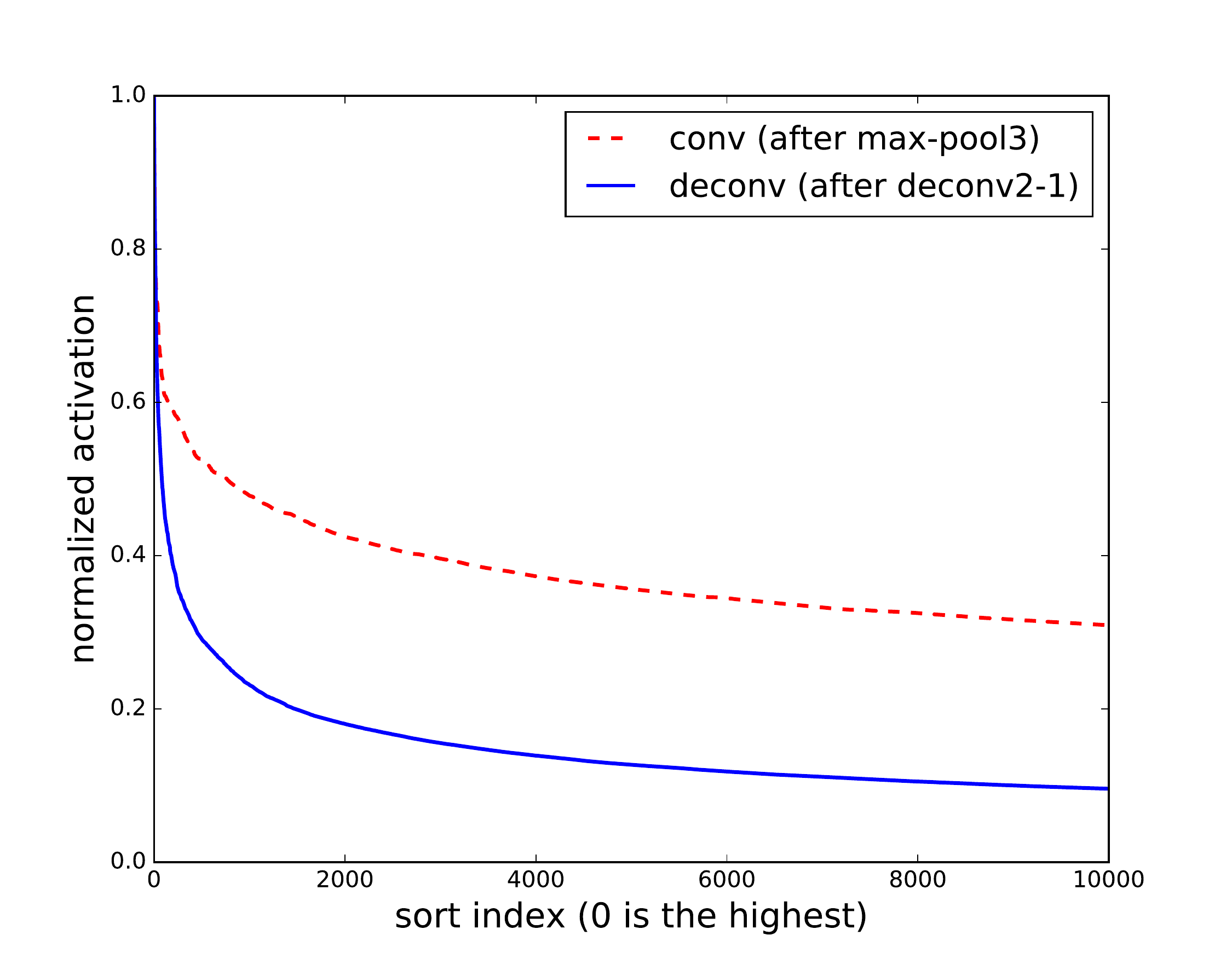}
\caption{Distribution of top most 10,000 activations in feature maps generated from convolution and corresponding deconvolution layers. Tied deconvolution layers are stacked on \textit{VGG-16}~\cite{b3_simonyan2014cls_vgg} convolution layers, and the outputs of the \textit{3}$^{rd}$ pooling operation and its corresponding deconvolution counterpart are depicted as red-dashed (\textit{conv}) and blue-solid (\textit{deconv}) lines, respectively. Top most activations are more discriminative in \textit{deconv} compared to \textit{conv} in terms of relative difference between the left most and the right most activations.}
\vskip -0.2in
\label{fig2:FP_reduction}
\end{center}
\end{figure} 

\subsection{Analysis}
\begin{figure}[t]
\begin{center}
\includegraphics[width=\columnwidth]{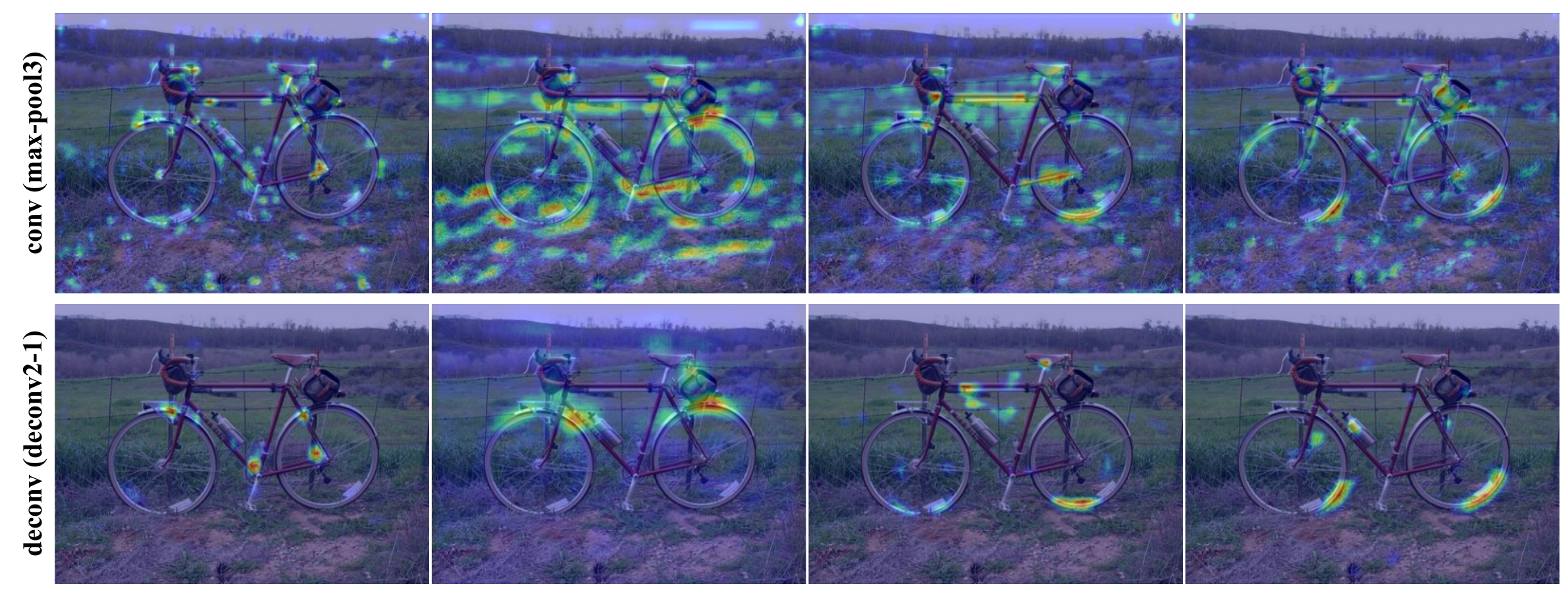}
\caption{Examples of the false positive reduction effect (best viewed in color). Top and bottom rows show some individual feature maps of \textit{conv} and their corresponding feature maps of \textit{deconv}. Both are the same as \textit{conv} and \textit{deconv} in Fig.~\ref{fig2:FP_reduction}. Feature maps extracted from a convolution layer are more noisy compared to the feature maps from the corresponding deconvolution layer while the most discriminative parts in both cases are similarly maintained.} 
\vskip -0.2in
\label{fig3:FP_reduction_ex}
\end{center}
\end{figure}

Fig.~\ref{fig2:FP_reduction} shows the most discriminative 10,000 activations in a pair of convolution and deconvolution layers. Based on a \textit{VGG-16}~\cite{b3_simonyan2014cls_vgg} pre-trained network except for the last fully-connected layers, deconvolution layers with tied weights are stacked on it (details of the network architecture and experimental set-up will be explained in Section 4). The red-dashed line describes the normalized activations of the feature map generated from a specific convolution layer (max-pool3 in the \textit{3}$^{rd}$ convolution layer), and the blue-solid line depicts that of the corresponding deconvolution layer. The difference between the left most and the right most activations in \textit{conv} is smaller than the difference of \textit{deconv} as shown in this figure. It means that the top most activations in \textit{deconv} are more discriminative in terms of their relative values to the rest activations. Fig.~\ref{fig3:FP_reduction_ex} shows four individual feature maps selected from a blob of feature maps extracted by the \textit{3}$^{rd}$ convolution layer (upper) and their matched feature maps of corresponding deconvolution layer (lower). The feature maps from the deconvolution layer are less noisy than the corresponding feature maps of the convolution layer while the most discriminative parts are mostly maintained. So, we use the feature maps generated by multiple deconvolution layers in order to build a rich and discriminative feature set which is robust against false positives.

\section{Experiment} 
We examine the proposed methodology on two different datasets; chest X-rays (CXRs) for lesion segmentation in Tuberculosis (TB) screening and natural images for object segmentation. Description of the datasets used for training, validation, and test will be presented in Section 4.1, and details of the implementation will be explained in 4.2. Section 4.3 shows the experimental results on both datasets as well as interpretation on the effect of the proposed methodology. 

\subsection{Dataset}
\noindent\textbf{TB}~~Three different de-identified CXR datasets, namely KIT (digitized by \textit{digital radiography}; DR-type), MC (digitized by \textit{computed radiography}; CR-type), and Shenzhen (DR-type) sets, are used in this study. KIT set (total 10,848 DICOM images) consists of 7,020 normal and 3,828 abnormal (TB) cases from Korean Institute of Tuberculosis (KIT) under Korean National Tuberculosis Association, South Korea. MC set has 80 normal and 58 abnormal cases from National Library of Medicine, National Institutes of Health, Bethedsda, MD, USA (Montgomery County), and Shenzhen set consists of 326 normal and 336 abnormal cases from Shenzhen No.3 People's Hospital, Guangdong Medical College, Shenzhen, China. 80\% and 20\% of CXRs in KIT set are used for training and validation. All the CXRs in KIT set only have image-level abnormality information. MC and Shenzhen sets, available for the research purpose~\cite{h1_Jaeger2013TB,h2_Jaeger2014TB,h3_Candemir2014TB}, are used for performance evaluation. We obtain lesion annotations of abnormal CXRs on both datasets from a TB clinician in order to measure the segmentation performance quantitatively.

\vskip 0.1in
\noindent\textbf{Object segmentation}~~PASCAL VOC classification dataset (20 classes)~\cite{i1_everingham2010voc} is used for training and validation. Among total 11,540 images in the classification dataset including 5,717 and 5,823 images for training and validation, we removed 2,273 images overlapped with the PASCAL VOC 2012 segmentation dataset, and also discarded 802 images with `ambiguous' labels. 80\% and 20\% of the remaining 8,465 images (only with image-level labels) are used for training and validation, respectively. We removed all the overlapped images between classification and segmentation datasets from our training set in order to evaluate the proposed method on conservative but exact experimental environment, so the size of training set is relatively small compared to previous works~\cite{e1_Pinheiro2015seg_weakly,e2_pathak2014seg_weakly,e3_pathak2015seg_weakly}. PASCAL VOC 2012 segmentation dataset has 1,464 and 1,449 images for training and validation, so we used 1,449 validation images for performance evaluation and comparison.

\setlength{\tabcolsep}{4pt}
\begin{table}[t]
\caption{Layer component of the proposed framework for TB screening. Convolutional filter weight of deconv(\textit{k}) is tied with that of conv(\textit{6-k}). (Number of filters, kernel size, stride) for conv or deconv, and (kernel size, stride) for pool or unpool.}
\label{table1:architecture_TB}
\def\arraystretch{1.5}
\begin{center}\scriptsize
	\begin{tabular}{ l | c c c c c }
	\hline\hline
				& conv1 		& conv2 		& conv3 		& conv4 		& conv5 		\\
	\hline
	conv		& 96 11x11 1	& 256 5x5 1 	& 384 3x3 1 	& 384 3x3 1 	& 256 3x3 1	\\
	pool		& 2x2 2			& 2x2 2			& 2x2 2			& 2x2 2			& 2x2 2			\\
	\hline
				& deconv1 	& deconv2 	& deconv3 	& deconv4 	& deconv5 	\\
	\hline
	unpool	& 2x2 2			& 2x2 2			& 2x2 2			& -				& -				\\
	deconv	& 384 3x3 1	& 384 3x3 1 	& 256 3x3 1 	& - 				& -				\\
	\hline\hline
	\end{tabular}
\end{center}
\end{table}
\setlength{\tabcolsep}{1.4pt}

\setlength{\tabcolsep}{4pt}
\begin{table}[t]
\caption{Layer component of the proposed framework for object segmentation. Convolutional filter weight of deconv(\textit{k})-\textit{i} is tied with that of conv(\textit{6-k})-\textit{i}. (Number of filters, kernel size, stride) for c\textit{(k)} or d\textit{(k)}, and (kernel size, stride) for pool or unpool.}
\label{table2:architecture_VOC}
\def\arraystretch{1.5}
\begin{center}\scriptsize
	\begin{tabular}{ l | c c c c c }
	\hline\hline
						& conv1 		& conv2 		& conv3 		& conv4 		& conv5 		\\
	\hline
	c\textit{(k)}-1	& 64 3x3 1		& 128 3x3 1 	& 256 3x3 1 	& 512 3x3 1 	& 512 3x3 1	\\
	c\textit{(k)}-2	& 64 3x3 1		& 128 3x3 1 	& 256 3x3 1 	& 512 3x3 1 	& 512 3x3 1	\\
	c\textit{(k)}-3	& -				& - 				& 256 3x3 1 	& 512 3x3 1 	& 512 3x3 1	\\
	pool				& 2x2 2			& 2x2 2			& 2x2 2			& 2x2 2			& 2x2 2			\\
	\hline
						& deconv1 	& deconv2 	& deconv3 	& deconv4 	& deconv5 	\\
	\hline
	unpool			& 2x2 2			& 2x2 2			& 2x2 2			& -				& -				\\
	d\textit{(k)}-3	& 512 3x3 1	& 512 3x3 1 	& 256 3x3 1 	& - 				& -				\\
	d\textit{(k)}-2	& 512 3x3 1	& 512 3x3 1 	& 256 3x3 1 	& - 				& -				\\
	d\textit{(k)}-1	& 512 3x3 1	& 256 3x3 1 	& 128 3x3 1 	& - 				& -				\\
	\hline\hline
	\end{tabular}
\end{center}
\vskip -0.2in
\end{table}
\setlength{\tabcolsep}{1.4pt}

\subsection{Experimental Set-up}
We train a CNN for TB from scratch, since we do not have any pre-trained models for CXRs. While, \textit{VGG-16}~\cite{b3_simonyan2014cls_vgg} pre-trained on the ImageNet classification dataset~\cite{b1_Deng2009imagenet} is used as a base CNN for the PASCAL VOC. Details of layer components are summarized in Table~\ref{table1:architecture_TB} and \ref{table2:architecture_VOC}. 

\vskip 0.1in
\noindent\textbf{TB}~~320$\times$320 images are used for training. All the CXRs are resized to 340$\times$340 and randomly cropped from the resized one. We cropped only 20 pixels away, since lesions in abnormal CXRs should be retained after cropping. Any additional augmentations (except for horizontal mirroring) allowable for lesion deformation are not adopted. End-to-end training on the proposed framework was not successful, because the KIT set used for training is too small (8.7k images for training). So, we train the model layer-by-layer after feature learning from a classification network. We first build and train a classification network using two layers of 2048 fully-connected nodes with an output softmax layer stacked on five convolution layers described in Table~\ref{table1:architecture_TB} (0.01 initial leaning rate is decayed by half in every 20 epochs until 50 epochs). After that, five trained convolution layers with an additional convolution operator (with randomly initialized weights) for class-specific activation maps (bottom of Fig.~\ref{fig1:overall_arch}) are trained accordingly (0.002 initial learning rate is decayed by half in every 20 epochs until 50 epochs). This is our baseline. From the baseline, each deconvolution layer described in Table~\ref{table1:architecture_TB} is stacked and trained in a layer-by-layer manner. For each deconvolution layer, learning rate 0.001 is used until 10 epochs without decaying.

\vskip 0.1in
\noindent\textbf{Object segmentation}~~224$\times$224 images are randomly cropped from 256$\times$256 resized images. We only use random cropping and mirroring for dataset augmentation in order to purely focus on advantages of the proposed methodology. We stack deconvolution layers on the \textit{VGG-16} base net except for the fully-connected layers, and perform end-to-end training with learning rate 0.001 without decaying until 10 epochs. In this case, convolutional weights except for the weights tied with deconvolution layers are fixed during training. 

\vskip 0.1in
Both networks are optimized with stochastic gradient descent (SGD) with a momentum 0.9. Minibatch size is 16, and similarity parameter \textit{s} in Eq~(\ref{eq7:LSE}) is 5. All the experiments are performed using \textit{theano}~\cite{i2_bastien2012theano}.

\setlength{\tabcolsep}{4pt}
\begin{table}[t]
\caption{Lesion segmentation results (IoU in \%) of TB screening in two public CXRs.}
\label{table3:IoU_TB}
\def\arraystretch{1.5}
\begin{center}\scriptsize
	\begin{tabular}{@{\extracolsep{\stretch{1}}}*{5}{ l | c c c c }@{}}
	\hline\hline
								& w/o deconv	& \multicolumn{3}{c}{w/ deconv}			\\
								& (baseline)	& stage1 		& stage2 		& stage3	\\
	\hline
	MC (US)				& 19.73			& 21.04			& 21.33			& 21.61 	\\
	Shenzhen (CHN)	& 17.39			& 20.02			& 21.49			& 24.61 	\\
	\hline\hline
	\end{tabular}
\end{center}
\vskip -0.2in
\end{table}
\setlength{\tabcolsep}{1.4pt}

\subsection{Quantitative Results}
We use intersection-over-union (IoU), a common assessment metric for semantic segmentation~\cite{i1_everingham2010voc}, for evaluation. 

\begin{figure*}
\begin{center}
\includegraphics[scale=0.48]{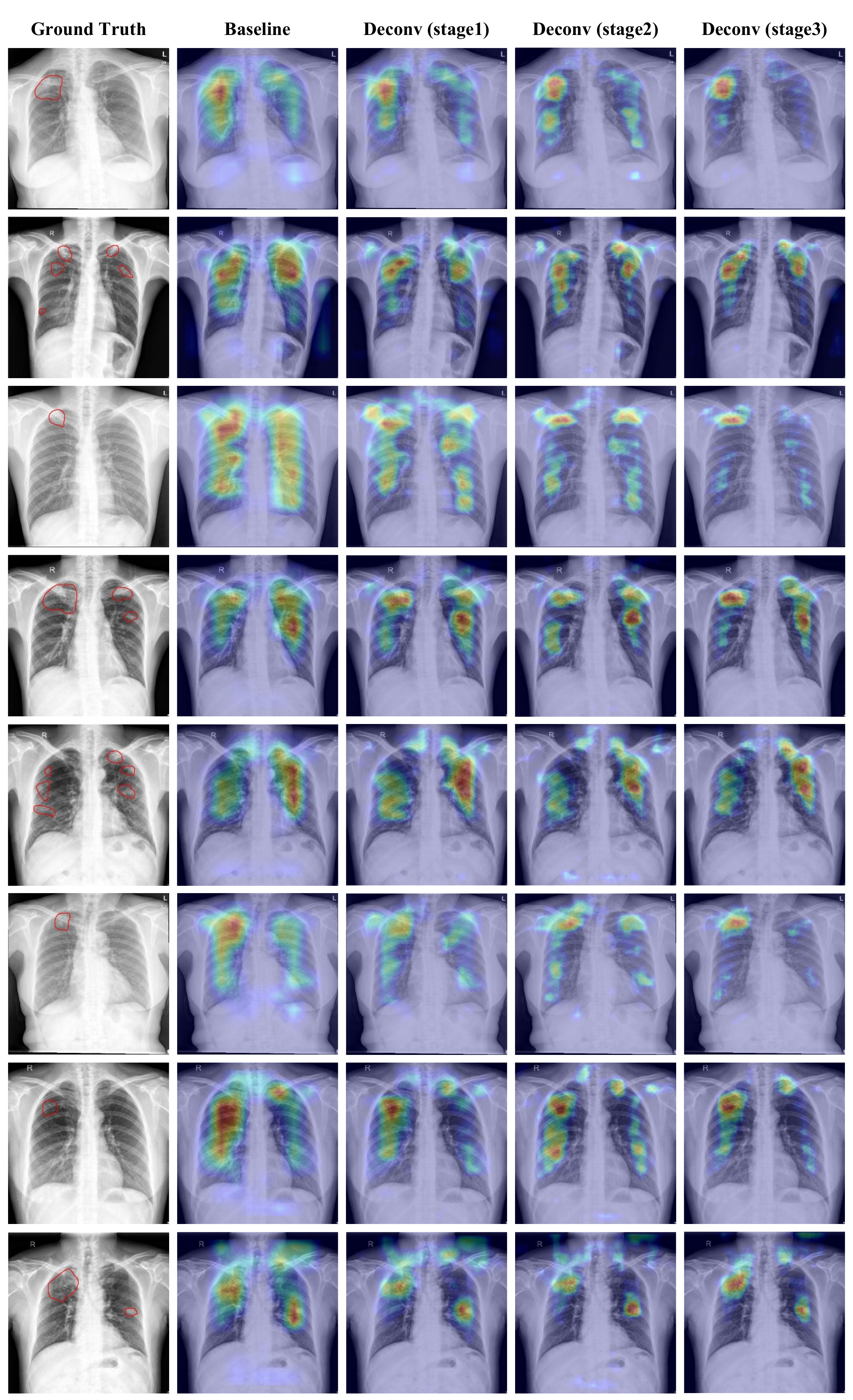}
\vskip -0.05in
\caption{Examples of class-specific output activation maps on CXRs (best viewed in color). False positives are reduced as deconvolution stages proceed while the detailed shape is gradually restored regardless of lesion scales.}
\label{fig4:results_TB}
\end{center}
\end{figure*}

\vskip 0.1in
\noindent\textbf{TB}~~Lesion segmentation results on MC and Shenzhen sets are summarized in Table~\ref{table3:IoU_TB}. The localization network in~\cite{e1_Pinheiro2015seg_weakly} is our baseline (\textit{log-sum-exp} per-map pooling) which is trained based on pre-trained features from a classification network using KIT set (DR-type). The proposed method achieves 9.6\% and 41.5\% of performance improvements in MC (CR-type) and Shenzhen (DR-type) sets compared to the baseline. \textit{Stage N} in Table~\ref{table3:IoU_TB} means that the proposed framework includes deconvolution layers stacked until deconv-\textit{N}. \textit{Stages} 1, 2, and 3 are trained layer-by-layer in order. Fig.~\ref{fig4:results_TB} shows examples of lesion segmentation results. Class-specific activation map is normalized and depicted as heat-map on an input CXR. The proposed method gradually discriminates details of lesions (e.g., the \textit{2}$^{nd}$ and \textit{5}$^{th}$ examples from the top) while eliminating false positives as stage proceeds. In some cases, uncertain lesions become clear as deconvolution layers are going deeper, e.g., unclear lesion on left top of the \textit{4}$^{th}$ example becomes more discriminative as stages proceed.

\setlength{\tabcolsep}{0.55pt}
\begin{table*}[t]
\caption{Object segmentation results (IoU in \%) of the PASCAL VOC 2012 validation set. Previous weakly-supervised approaches for semantic segmentation are FC-MIL~\cite{e2_pathak2014seg_weakly}, GP-LSE~\cite{e1_Pinheiro2015seg_weakly}, and C-CNN~\cite{e3_pathak2015seg_weakly}.}
\label{table4:mIoU_VOC}
\def\arraystretch{1.5}
\begin{center}\tiny
	\begin{tabular*}{\textwidth}{@{\extracolsep{\stretch{1}}}*{23}{ l c c c c c c c c c c c c c c c c c c c c c c }@{}}
	\hline\hline
					& bg	& aero & bike & bird & boat & btl & bus & car & cat & chr & cow & tbl & dog & hrs & mbk & per & plnt & shp & sofa & trn & tv & mIoU		\\
	\hline
	baseline	& 67.6 & 30.4 & 27.9 & 23.6 & 35.0 & 15.1 & 22.0 & 27.0 & 41.2 & 15.9 & 15.5 & 14.3 & 30.2 & 23.2 & 31.8 & 35.0 & 25.3 & 26.9 & 20.4 & 16.6 & 28.2 & 27.3 \\
	stage1	& 71.7 & 30.7 & 30.5 & 26.3 & 20.0 & 24.2 & 39.2 & 33.7 & 50.2 & 17.1 & 29.7 & \textbf{22.5} & \textbf{41.3} & 35.7 & \textbf{43.0} & 36.0 & 29.0 & \textbf{34.9} & 23.1 & 33.2 & 33.2 & \textbf{33.6}  \\
	stage2	& \textbf{73.1} & \textbf{32.2} & \textbf{30.7} & \textbf{30.0} & \textbf{21.6} & 24.4 & 37.0 & 33.3 & \textbf{51.5} & \textbf{18.6} & 24.1 & 17.1 & 38.8 & \textbf{36.9} & 40.6 & \textbf{41.0} & \textbf{32.1} & 28.4 & \textbf{24.2} & 33.2 & 33.4 & 33.4  \\
	stage3	& 70.4 & 28.1 & 27.6 & 22.1 & 14.7 & 22.0 & 35.2 & 32.2 & 44.0 & 15.4 & 24.2 & 12.9 & 36.0 & 32.7 & 41.5 & 33.4 & 26.5 & 28.7 & 21.0 & 32.7 & 33.1 & 30.2 \\
	\hline
	FC-MIL	& -       & -       & -       & -      & -       & -       & -       & -       & -      & -       & -       & -       & -      & -       & -       & -       & -       & -      & -       & -       & -       & 24.9 \\
	GP-LSE		& 37.0 & 10.4 & 12.4 & 10.8 &   5.3 &   5.7 & 25.2 & 21.1 & 25.2 &   4.8 & 21.5 &   8.6 & 29.1 & 25.1 & 23.6 & 25.5 & 12.0 & 28.4 &   8.9 & 22.0 & 11.6 & 17.8  \\
	C-CNN		& 66.3 & 24.6 & 17.2 & 24.3 & 19.5 & \textbf{34.4} & \textbf{45.6} & \textbf{44.3} & 44.7 & 14.4 & \textbf{33.8} & 21.4 & 40.8 & 31.6 & 42.8 & 39.1 & 28.8 & 33.2 & 21.5 & \textbf{37.4} & \textbf{34.4} & 33.3  \\	
	\hline\hline
	\end{tabular*}
\end{center}
\vskip -0.2in
\end{table*}
\setlength{\tabcolsep}{1.4pt}

\vskip 0.1in
\noindent\textbf{Object segmentation}~~Table~\ref{table4:mIoU_VOC} shows object segmentation results on a validation set in the PASCAL VOC 2012 segmentation dataset. Baseline includes the pre-trained \textit{VGG-16} except for the fully-connected layers with an additional convolution operator for class-specific map generation. \textit{Stages} 1, 2, and 3 are obtained independently through end-to-end training. 

Previous weakly-supervised semantic segmentation approaches, fully convolutional MIL (FC-MIL)~\cite{e2_pathak2014seg_weakly}, global pooling with \textit{log-sum-exp} (GP-LSE)~\cite{e1_Pinheiro2015seg_weakly}, and constrained CNN (C-CNN)~\cite{e3_pathak2015seg_weakly}, are also compared with the proposed method as described in Table~\ref{table4:mIoU_VOC}. All the comparison targets are purely based on a weakly-supervised setting without any types of pre/post processing or additional supervisions for fair comparison. Our method outperforms FC-MIL and GP-LSE. Performance difference between the baseline and GP-LSE (basic model of our baseline is the same with GP-LSE) comes from the difference between base CNNs (GP-LSE is based on \textit{Overfeat}~\cite{i3_sermanet2014overfeat}, while our baseline used \textit{VGG-16}~\cite{b3_simonyan2014cls_vgg}). 

\begin{figure}[t]
\begin{center}
\includegraphics[width=\textwidth]{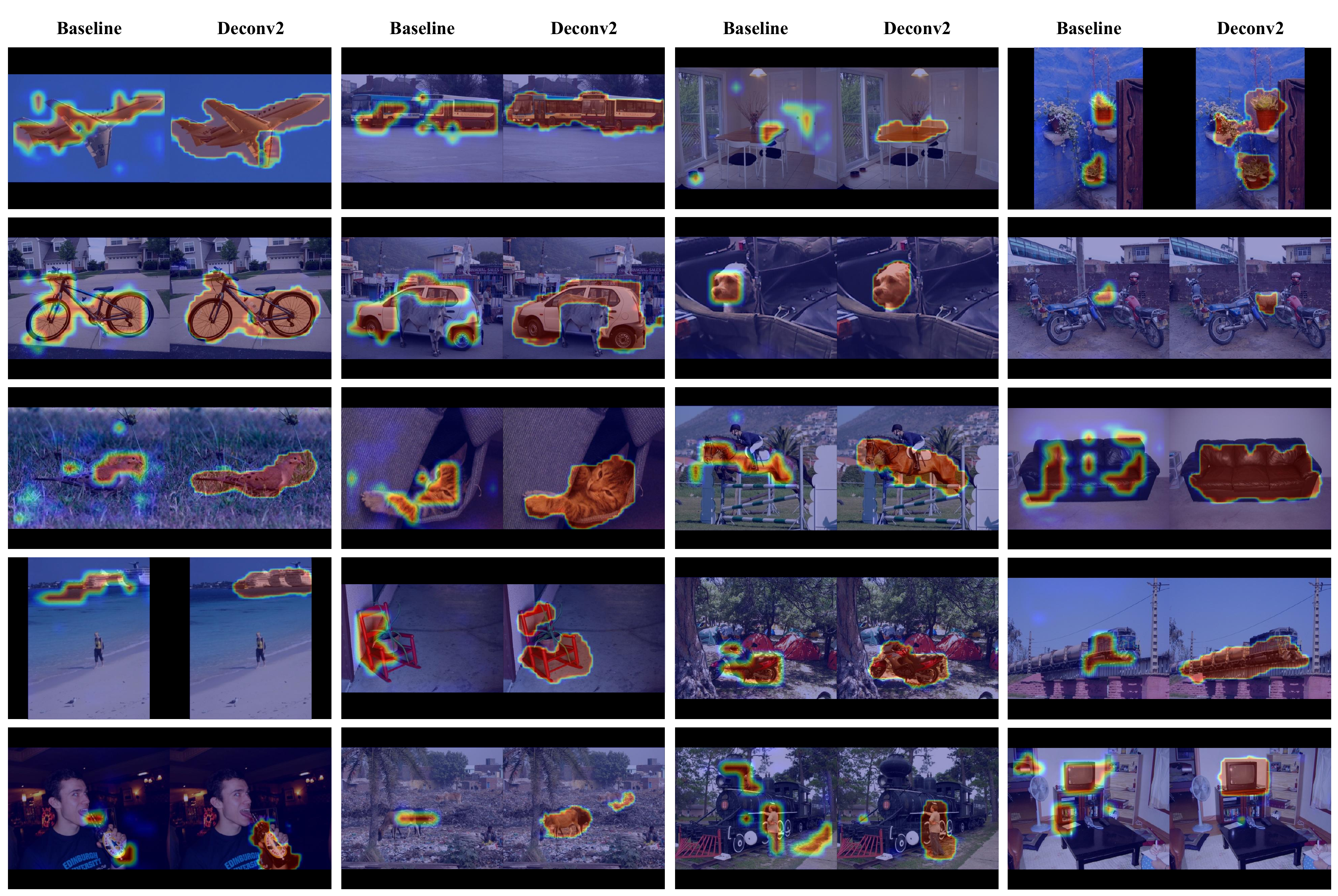}
\vskip -0.05in
\caption{Examples of class-specific output activation maps on the PASCAL VOC 2012 segmentation dataset (best viewed in color). Boundaries of ROIs become more delicate and false positives are reduced with the proposed framework.}
\vskip -0.2in
\label{fig5:results_VOC}
\end{center}
\end{figure}

C-CNN defines the target objective as a convex optimization problem as mentioned in Section 2. It necessarily requires additional optimization procedure to approximate a latent probability distribution, \textbf{P(x)}, of an output distribution of CNN, \textbf{Q(x)}, in every iteration. Although we used 8,465 images for training (6,772 for training and 1,693 for validation) under the conservative set-up for datasets (Section 4.2), our method shows slightly better performance compared to C-CNN (total 10,582 training images).
Fig.~\ref{fig5:results_VOC} shows class-specific activation maps of 20 objects in different classes. Normalized class-specific activation maps are overlaid on input images. As shown in this figure, detailed boundaries with reduced false positives are well described with the proposed method.

\section{Conclusion} 
In this work, we proposed a weakly-supervised semantic segmentation framework based on a tied deconvolutional neural network. A bunch of discriminative features with different abstraction levels along with different sizes of receptive fields are extracted in order to construct a rich feature set to be selectively used for ROI segmentation. Unpoolings help to reduce false positives while maintaining relevant discriminative features, and tied deconvolutions properly localize ROIs under weak supervision with a tight connection between encoding and decoding paths of the entire network. Two different datasets, medical images and natural images, are used for verifying applicability of the proposed method. Especially, the proposed method can be a practical solution for computer-aided-detection/diagnosis (CADe/CADx) in a medical imaging domain, since lesion annotation under domain-specific knowledges requires substantial expenses compared to annotating ROIs on natural images.


\bibliographystyle{splncs}
\bibliography{eccv2016_deconvnet_hekim}

\begin{thebibliography}{10}

\bibitem{b1_Deng2009imagenet}
Deng, J., Dong, W., Socher, R., Li, L.J., Li, K., Fei-Fei, L.:
\newblock Imagenet: A large-scale hierarchical image database.
\newblock In: Computer Vision and Pattern Recognition (CVPR). (2009)

\bibitem{b2_Krizhevsky2012cls_alexnet}
Krizhevsky, A., Sutskever, I., Hinton, G.E.:
\newblock Imagenet classification with deep convolutional neural networks.
\newblock In: Advances in Neural Information Processing Systems (NIPS). (2012)

\bibitem{b3_simonyan2014cls_vgg}
Simonyan, K., Zisserman, A.:
\newblock Very deep convolutional networks for large-scale image recognition.
\newblock In: International Conference on Learning Representations (ICLR).
  (2015)

\bibitem{a1_hinton2012speech}
Hinton, G., Deng, L., Yu, D., Dahl, G.E., Mohamed, A.r., Jaitly, N., Senior,
  A., Vanhoucke, V., Nguyen, P., Sainath, T.N.,  et~al.:
\newblock Deep neural networks for acoustic modeling in speech recognition: The
  shared views of four research groups.
\newblock Signal Processing Magazine, IEEE \textbf{29}(6) (2012)  82--97

\bibitem{a2_dahl2012speech}
Dahl, G.E., Yu, D., Deng, L., Acero, A.:
\newblock Context-dependent pre-trained deep neural networks for
  large-vocabulary speech recognition.
\newblock Audio, Speech, and Language Processing, IEEE Transactions on
  \textbf{20}(1) (2012)  30--42

\bibitem{a3_collobert2008nlp}
Collobert, R., Weston, J.:
\newblock A unified architecture for natural language processing: Deep neural
  networks with multitask learning.
\newblock In: International Conference on Machine Learning (ICML). (2008)

\bibitem{a4_cho2014nlp}
Cho, K., Gulcehre, B.v.M.C., Bahdanau, D., Schwenk, F.B.H., Bengio, Y.:
\newblock Learning phrase representations using rnn encoder--decoder for
  statistical machine translation.
\newblock In: Conference on Empirical Methods in Natural Language Processing
  (EMNLP). (2014)

\bibitem{b4_bell2015det_insideoutside}
Bell, S., Zitnick, C.L., Bala, K., Girshick, R.:
\newblock Inside-outside net: Detecting objects in context with skip pooling
  and recurrent neural networks.
\newblock arXiv preprint arXiv:1512.04143 (2015)

\bibitem{b5_Erhan2014det}
Erhan, D., Szegedy, C., Toshev, A., Anguelov, D.:
\newblock Scalable object detection using deep neural networks.
\newblock In: Computer Vision and Pattern Recognition (CVPR). (2014)

\bibitem{b6_Oquab2015det_free}
Oquab, M., Bottou, L., Laptev, I., Sivic, J.:
\newblock Is object localization for free?--weakly-supervised learning with
  convolutional neural networks.
\newblock In: Computer Vision and Pattern Recognition (CVPR). (2015)

\bibitem{c1_long2014seg_fully}
Long, J., Shelhamer, E., Darrell, T.:
\newblock Fully convolutional networks for semantic segmentation.
\newblock In: Computer Vision and Pattern Recognition (CVPR). (2015)

\bibitem{d2_hong2015seg_semi}
Hong, S., Noh, H., Han, B.:
\newblock Decoupled deep neural network for semi-supervised semantic
  segmentation.
\newblock In: Advances in Neural Information Processing Systems (NIPS). (2015)

\bibitem{d1_dai2015seg_semi}
Dai, J., He, K., Sun, J.:
\newblock Boxsup: Exploiting bounding boxes to supervise convolutional networks
  for semantic segmentation.
\newblock arXiv preprint arXiv:1503.01640 (2015)

\bibitem{d3_papandreou2015seg_semi}
Papandreou, G., Chen, L.C., Murphy, K., Yuille, A.L.:
\newblock Weakly-and semi-supervised learning of a dcnn for semantic image
  segmentation.
\newblock arXiv preprint arXiv:1502.02734 (2015)

\bibitem{e1_Pinheiro2015seg_weakly}
Pinheiro, P.O., Collobert, R.:
\newblock From image-level to pixel-level labeling with convolutional networks.
\newblock In: Computer Vision and Pattern Recognition (CVPR). (2015)

\bibitem{e3_pathak2015seg_weakly}
Pathak, D., Krahenbuhl, P., Darrell, T.:
\newblock Constrained convolutional neural networks for weakly supervised
  segmentation.
\newblock In: International Conference on Computer Vision (ICCV). (2015)

\bibitem{e4_hong2015seg_weakly}
Hong, S., Oh, J., Han, B., Lee, H.:
\newblock Learning transferrable knowledge for semantic segmentation with deep
  convolutional neural network.
\newblock arXiv preprint arXiv:1512.07928 (2015)

\bibitem{e5_zhang2015seg_weakly}
Zhang, W., Zeng, S., Wang, D., Xue, X.:
\newblock Weakly supervised semantic segmentation for social images.
\newblock In: Computer Vision and Pattern Recognition (CVPR). (2015)

\bibitem{c2_chen2014seg_fully}
Chen, L.C., Papandreou, G., Kokkinos, I., Murphy, K., Yuille, A.L.:
\newblock Semantic image segmentation with deep convolutional nets and fully
  connected crfs.
\newblock In: International Conference on Learning Representation (ICLR).
  (2015)

\bibitem{c3_hariharan2014seg_fully}
Hariharan, B., Arbel{\'a}ez, P., Girshick, R., Malik, J.:
\newblock Hypercolumns for object segmentation and fine-grained localization.
\newblock In: Computer Vision and Pattern Recognition (CVPR). (2015)

\bibitem{c4_hariharan2014seg_fully}
Hariharan, B., Arbel{\'a}ez, P., Girshick, R., Malik, J.:
\newblock Simultaneous detection and segmentation.
\newblock In: European Conference on Computer Vision (ECCV). (2014)

\bibitem{c5_mostajabi2014seg_fully}
Mostajabi, M., Yadollahpour, P., Shakhnarovich, G.:
\newblock Feedforward semantic segmentation with zoom-out features.
\newblock In: Computer Vision and Pattern Recognition (CVPR). (2015)

\bibitem{c6_noh2015seg_fully}
Noh, H., Hong, S., Han, B.:
\newblock Learning deconvolution network for semantic segmentation.
\newblock In: International Conference on Computer Vision (ICCV). (2015)

\bibitem{c7_dai2015seg_fully}
Dai, J., He, K., Sun, J.:
\newblock Instance-aware semantic segmentation via multi-task network cascades.
\newblock arXiv preprint arXiv:1512.04412 (2015)

\bibitem{c8_pinheiro2015seg_fully}
Pinheiro, P.O., Collobert, R., Dollar, P.:
\newblock Learning to segment object candidates.
\newblock In: Advances in Neural Information Processing Systems (NIPS). (2015)

\bibitem{g1_zeiler2011deconv}
Zeiler, M.D., Taylor, G.W., Fergus, R.:
\newblock Adaptive deconvolutional networks for mid and high level feature
  learning.
\newblock In: International Conference on Computer Vision (ICCV). (2011)

\bibitem{g2_zeiler2014deconv}
Zeiler, M.D., Fergus, R.:
\newblock Visualizing and understanding convolutional networks.
\newblock In: European Conference on Computer Vision (ECCV).
\newblock (2014)

\bibitem{g3_vincent2010sdae}
Vincent, P., Larochelle, H., Lajoie, I., Bengio, Y., Manzagol, P.A.:
\newblock Stacked denoising autoencoders: Learning useful representations in a
  deep network with a local denoising criterion.
\newblock The Journal of Machine Learning Research (JMLR) \textbf{11} (2010)
  3371--3408

\bibitem{i1_everingham2010voc}
Everingham, M., Van~Gool, L., Williams, C.K., Winn, J., Zisserman, A.:
\newblock The pascal visual object classes (voc) challenge.
\newblock International Journal of Computer Vision (IJCV) \textbf{88}(2) (2010)
   303--338

\bibitem{f1_maron1998mil}
Maron, O., Lozano-P{\'e}rez, T.:
\newblock A framework for multiple-instance learning.
\newblock In: Advances in Neural Information Processing Systems (NIPS). (1998)

\bibitem{f2_Andrews2002mil}
Andrews, S., Tsochantaridis, I., Hofmann, T.:
\newblock Support vector machines for multiple-instance learning.
\newblock In: Advances in Neural Information Processing Systems (NIPS). (2002)

\bibitem{h1_Jaeger2013TB}
Jaeger, S., Karargyris, A., Candemir, S., Siegelman, J., Folio, L., Antani, S.,
  Thoma, G.:
\newblock Automatic screening for tuberculosis in chest radiographs: a survey.
\newblock Quantitative Imaging in Medicine and Surgery \textbf{3}(2) (2013)
  89--99

\bibitem{h2_Jaeger2014TB}
Jaeger, S., Karargyris, A., Candemir, S., Folio, L., Siegelman, J., Callaghan,
  F., Xue, Z., Palaniappan, K., Singh, R.K., Antani, S.,  et~al.:
\newblock Automatic tuberculosis screening using chest radiographs.
\newblock IEEE Transactions on Medical Imaging \textbf{33}(2) (2014)  233--245

\bibitem{h3_Candemir2014TB}
Candemir, S., Jaeger, S., Palaniappan, K., Musco, J.P., Singh, R.K., Xue, Z.,
  Karargyris, A., Antani, S., Thoma, G., McDonald, C.J.:
\newblock Lung segmentation in chest radiographs using anatomical atlases with
  nonrigid registration.
\newblock IEEE Transactions on Medical Imaging \textbf{33}(2) (2014)  577--590

\bibitem{e2_pathak2014seg_weakly}
Pathak, D., Shelhamer, E., Long, J., Darrell, T.:
\newblock Fully convolutional multi-class multiple instance learning.
\newblock arXiv preprint arXiv:1412.7144 (2014)

\bibitem{i2_bastien2012theano}
Bastien, F., Lamblin, P., Pascanu, R., Bergstra, J., Goodfellow, I.J.,
  Bergeron, A., Bouchard, N., Bengio, Y.:
\newblock Theano: new features and speed improvements.
\newblock In: Advances in Neural Information Processing Systems (NIPS) Deep
  Learning and Unsupervised Feature Learning Workshop. (2012)

\bibitem{i3_sermanet2014overfeat}
Sermanet, P., Eigen, D., Zhang, X., Mathieu, M., Fergus, R., LeCun, Y.:
\newblock Overfeat: Integrated recognition, localization and detection using
  convolutional networks.
\newblock In: International Conference on Learning Representation (ICLR).
  (2014)

\end{thebibliography}
\end{document}